\documentclass[letterpaper, 10 pt, conference]{ieeeconf}  

\IEEEoverridecommandlockouts                              
\usepackage[utf8]{inputenc}
\usepackage[T1]{fontenc}

\usepackage{cite}
\usepackage{graphicx}
\usepackage{xcolor}
\usepackage{balance} 
\usepackage{tabularx}

\title{\LARGE \bf
Magic in Human-Robot Interaction (HRI)*
}

\author{Martin Cooney and Alexey Vinel$^{1}$
\thanks{*We gratefully acknowledge support from the Swedish
Knowledge Foundation for the ``Safety of Connected
Intelligent Vehicles in Smart Cities – SafeSmart'' project
(2019–2023), the Swedish Innovation Agency (VINNOVA)
for the ``Emergency Vehicle Traffic Light Pre-emption in
Cities – EPIC'' project (2020–2022), and the ELLIIT Strategic
Research Network.}
\thanks{$^{1}$M. Cooney and A. Vinel are with the School of Information Technology, Halmstad University, Halmstad, Sweden
        {\tt\small martin.daniel.cooney at gmail.com, alexey.vinel at hh.se}}%
}

\begin{document}

\maketitle
\thispagestyle{empty}
\pagestyle{empty}

\begin{abstract}
"Magic" is referred to here and there in the robotics literature, from "magical moments" afforded by a mobile bubble machine, to "spells" intended to entertain and motivate children--but what exactly could this concept mean for designers?
Here, we present (1) some theoretical discussion on how magic could inform interaction designs based on reviewing the literature, followed by (2) a practical description of using such ideas to develop a simplified prototype, which received an award in an international robot magic competition.
Although this topic can be considered unusual and some negative connotations exist (e.g., unrealistic thinking can be referred to as magical), our results seem to suggest that magic, in the experiential, supernatural, and illusory senses of the term, could be useful to consider in various robot design contexts, also for artifacts like home assistants and autonomous vehicles--thus, inviting further discussion and exploration.
\end{abstract}

\section{INTRODUCTION}

\begin{figure}[t]
\includegraphics[width=.5\textwidth]{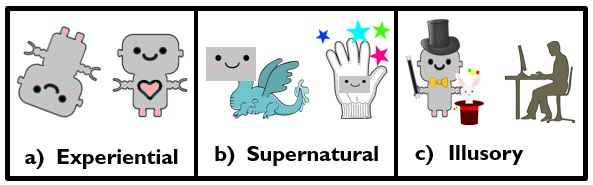}
\caption{Basic Concept: Robots can incorporate "magic" in its (a) experiential sense (e.g., in play and affection), (b) supernatural sense (e.g., in emulating fantastical creatures and objects), and (c) illusory sense (e.g., in tricks and Wizard-of-Oz scenarios).} \label{fig_basic_concept}
\end{figure}

This paper explores the concept of "magic" in robotics, as depicted in Fig.~\ref{fig_basic_concept}.

Our basic motivation concerns a bias in past robotics research to focus on utilitarian tasks and fundamental concerns such as efficacy and safety--that is now shifting to also explore the broader picture of how to support human aspirations and well-being through positive user experiences~\cite{luria2020destruction}.
For example, as robotic artifacts like Home Assistants (HAs) and Autonomous Vehicles (AVs) become increasingly prevalent, we might want to know how to get patients to relax when caregivers are busy or how to alleviate boredom in drivers who no longer have to drive. 
One potential key for unlocking positive experiences, that is receiving some attention in human-computer interaction and media science, is \emph{magic}: e.g., "The next big step for those working in multimedia is to develop and implement experiences that, until this point, were considered the domain of magic and magicians"~\cite{boll2008magic}.

But what exactly is meant by such terms?
Tom Robbins's mused that "using words to describe magic is like using a screwdriver to cut roast beef";\footnote{everydaypower.com/magic-quotes/}
nonetheless, such concepts could be difficult to "swallow" whole, so some definitions are carved out below, to aid in analysis and ideation:
A "robot" is a computing artifact that can sense in an intelligent or human-like way to interact, and move people or objects.
"HA" here is intended to refer to domestic robots--socially assistive robots, social robots, service robots, assistive robots, toy robots, and personal robots in homes, like care robots or vacuum cleaner robots--as well as automative technologies like smart displays and speakers.
"AV" refers to self-driving cars, driver-less cars, or robotic cars that can sense their environment and transport people or goods with little or no human control.
"Magic" here means "fantastic, special, exciting, wonderful, easy, and liked; resembling supernatural powers in beings or objects; or containing some implausible trick."\footnote{dictionary.cambridge.org/dictionary/english/magic}$^,$\footnote{thefreedictionary.com/magic}

In more detail, "magic" is an ancient word, originally referring to priesthood and "abilities"; relating to science, religion, and art; and possessing both positive and negative connotations due to various usages of the term by different groups over centuries--which here is described in a simplified manner in three main senses, as follows:
\begin{enumerate}
  \item \emph{Experiential}: Magical experiences are "those moments which are deemed to be both moving and memorable and thus are those that people really value"~\cite{reid2005magic}.
  \item \emph{Supernatural}: Part of legends, fantasies, and dreams,
  magic in a positive sense relates to blessings; cultural behaviors like religious rites can be intended to connect people to something greater, mystical, numinous, sacred, or spiritual--providing catharsis, stimulation, wish-fulfillment, or escape (e.g., Marett~\cite{marett1904spell} and Freud~\cite{ferenczi1925stages}), or self-actualization and closeness with nature (e.g., the Wicca movement). Magic in a negative sense relating to curses has also been used to express \emph{alterity}, power and superiority (e.g., in reference to illogical "magical thinking", or "primitive" peoples that believe in witches). 
  \item \emph{Illusory}: Positive examples include illusions intended to entertain or provoke thought, whereas negative examples include deceit by charlatans (and possibly even false accusations of witchcraft).
\end{enumerate}

These three categories are not fully independent--all incorporate some fantastical feeling--but exhibit differences: The first sense of the term can have nothing to do with supernatural appearance or intended tricks, the second involves supernatural appearance without intended tricks, and the third involves both supernatural appearance and intended tricks. Also, these categories are intended here to group related concepts--to help gain insight into subjective feelings and experiences--without implying anything about the physical or metaphysical nature of phenomena (i.e., our focus is on assertions like "this feels magical" rather than "this is magical").

Based on these definitions, we apply a speculative prototyping approach to investigate magic in Human-Robot Interaction (HRI): First, we review the robotics literature, bringing together various work along the three themes we identified, in Section~\ref{section:related-work}, which is then used to derive some speculative examples of how the concept could be incorporated into robot designs in Section~\ref{section:ideation}. Next, we describe an implementation of some of these concepts in a physical robot in Section~\ref{section:implementation}, concluding with some brief additional discussion in Section~\ref{section:discussion}.

\section{RELATED WORK}
\label{section:related-work}

After an initial check that did not reveal a review about magic in HRI, a more complete search was conducted using ACM, IEEE Xplore, and Google Scholar with the search phrase "(robot OR HRI OR "human-robot interaction" OR "home assistant" OR "autonomous vehicle") AND (magic OR fantasy)". ("Fantasy" was included due to its similarities with "magic".) The top fifty results for each database were checked without any restrictions on the year of publication, kind of paper, or target demographic, then duplicates and unrelated papers were removed. Some hand-picking of useful references from the bibliographies of the initially selected papers was also conducted. Thus, from 150 papers searched, approximately 50 papers were selected for inclusion below, which were sorted roughly into three categories aligned with our definition of magic from the previous section.

In doing so, we noticed that the words "magic" or "fantasy" appeared in many papers but often only sparsely (sometimes only once). Thus, roboticists seem to feel that these concepts bear some relevance, but tend to not consider them as a main component in designs.
Related to this, a broad range of topics was observed, including for example, discussion about recreational dolls, fictional wizards, and special effects in movies; at the risk of veering off-topic, we included many of these papers for the sake of completeness and in the hope that they might lead to useful discussion.
Also, surprisingly to us, despite the negative connotations described previously, we did not see any negative usages of the word magic in the selected papers--although it was sometimes used in unrelated product names and acronyms (e.g., in regard to the Multi Autonomous Ground-robotic International Challenge). By contrast, "fantasy" was often used in a negative sense to mean the opposite of "reality". Also, few papers related to HAs and AVs, suggesting that a gap might exist, or that other terms could also be considered in future searches.

\subsection{Experiential Magic}

While we believe that typical design guidelines can be followed to achieve good experiences in various contexts, our search turned up some papers that related to magic as a special feeling in two "hedonic" contexts in particular, play and affection.

\subsubsection{Play}
Magic has long been associated with play. 
Huizinga in his pioneering study used the term "magic circle" to describe a space for play outside of daily concerns~\cite{huizinga1950homo}.
Playful consideration of silly or "un-useless" magical machines has also been proposed as a way to explore problem spaces while making room for so-called "naïve, fragile fictions"; the idea thereby is to deter "solutionism", a bias toward finding solutions at all costs, even when a corresponding problem does not exist, or when some related problem must be "dumbed-down" to fit~\cite{blythe2016anti}.
Such "Ludic" considerations have been reflected in the design of various robots, including BubbleBot, which enabled "magic moments" by blowing bubbles while moving around people, eliciting enjoyment and sometimes also interactions between strangers~\cite{lee2020ludic}.

Furthermore, such magic is not considered a trivial add-on; one pediatric study reported that "it is the playful magic of the interaction that contributes to pain management and anxiety reduction"~\cite{mott2021design}.
Moreover, of various kinds of play that exist, open-ended play has been described as particularly useful for supporting interactions characterized by fantasy, imagination, and make-believe~\cite{alves2021children}.
As well, a special "magic" in novel haptic interfaces has been reported that might arise from unexpectedness and the pleasure of physical control~\cite{xu2008search}.
Playfulness can also been seen in responses from HAs,\footnote{cheatsheet.com/gear-style/20-questions-to-ask-siri-for-a-hilarious-response.html} as well as attempts to break down the "fourth wall" via commercials that trigger HAs.\footnote{theguardian.com/business/2017/apr/12/burger-king-ok-google-commercial}
Additionally, vehicles have been described as toys for adults that facilitate play~\cite{tranter1996reclaiming}, as well as dreams, freedom, and the communication of individuality~\cite{ingrassia2012engines}.

\subsubsection{Affection}
Another area in which robots and fantasies were described was in regard to love.
Despite taboos related to sex~\cite{earle2021involving} and a risk of wandering off-topic, since most robots are not intended for this purpose, these papers are included here for completeness, and due to love being perceived as an important topic that is "magical, exceptional, powerful, and transformative"; e.g., in children's movies, some characters have been depicted as wrapped in "magical swirls of sparks, leaves, or fireworks as they stare into each other’s eyes" accompanied by momentous music coming from nowhere, flowers, candles, fire, and beautiful nature (e.g., fireflies, butterflies, sunsets, and powerful wind)~\cite{martin2009hetero}.
Various studies have reported people affectionately bonding with robots that could act as partners or colleagues, in giving them names, and holding funerals and honoring them with medals when they break~\cite{nyholm2020can}.
A basis for romantic feelings toward robots also exists in relation to \emph{objectophilia} and \emph{agalmatophilia}, which refer to love of objects and dolls, respectively; one study reported that such robots will likely not be just used for sex, but rather that their owners might enjoy their companionship and spent time fantasizing about them~\cite{gonzalez2020human}.
Another study confirmed that a relationship with such a robot might be viewed less negatively than infidelity with a human partner--however, perceptions could sour in the future as robots become less like objects and more like fantastic playmates, that could even rival humans in some qualities~\cite{rothstein2021perceptions}.

\subsubsection{Potential concerns}
Continuing along a similar line, experiential magic afforded by play and affection might not always be beneficial.
For example, one study, which investigated if robots could be used to detect maltreatment through conversing with children, warned that the risk of false disclosures could be increased if playful robots were to encourage children's fantasies~\cite{henkel2017robot}; thus, serious tasks might require seriousness.
Also, negative effects of love robots could include setting unrealistic expectations and potentially promoting unhealthy behaviors~\cite{bartneck2021psychological}.

\subsection{Supernatural Magic}

Some work has also been done to emulate or recreate magic creatures, objects, settings, and behaviors; we note that the focus therein is not on a trick performed by a magician, but rather on depicting something fantastical.
Such robots could be useful, in line with Arthur C. Clarke's third law--that "any sufficiently advanced technology is indistinguishable from magic"--since belief in magic, as in religion, can increase well-being and happiness~\cite{ullrich2016social}.

\subsubsection{Creatures}

In addition to various movie props, robots with fantastical appearances have been built for research, which resemble implausible animals (gremlins, abstracted birds, living dinosaurs, and dragons) or spirits (skeletons, elves, wisps and heavenly beings):
For example, Leonardo, a furry robot reminiscent of sphinx cats or the magical creatures in the movie Gremlins (1984),\footnote{en.wikipedia.org/wiki/Gremlins} was referred to as "fantastical" by Singh et al. (the authors contrast this design with their own work, which leverages people's familiarity with dogs to communicate emotions via a tail~\cite{singh2013dog}). 
Described as "chick-like", the small rhythmic robot Keepon also has an abstracted and fantastical appearance, somewhat like a snowman or stack of Japanese "dango" dumplings~\cite{kozima2009keepon}.
Likewise, PLEO, which recently enabled positive interactions with hospitalized children, has the appearance of a small "living" dinosaur~\cite{moerman2021using}, and robots have even been built to resemble dragons~\cite{zhao2018design}.

More "spirit-like" robots have also been built, such as \emph{anthropomimetic} skeletons~\cite{diamond2012anthropomimetic} and elf-like phones~\cite{sumioka2012motion}.
Additionally, one robot's eye movements, trailing in a ghostly manner, were named after the Will-o'-the-wisp, a magic spirit said to have misled travellers at night~\cite{matsumaru2003robot}.
Another interesting study has advocated the use of "theomorphic" robots to represent divine, god-like qualities, highlighting that the hurdle to acceptance could be low, since people are used to praying without receiving a response~\cite{trovato2018design}; some examples of such robots include Mindar, a representation of Kannon, the Buddhist Goddess of Mercy, and SanTO, a saint-like robot~\cite{balle2020robots}. 
In our own work, we also built some fantastical prototypes, such as an "angelic" flying humanoid robot~\cite{cooney2012designing}, as well as a "vampire" that harvested a user's energy, and a "shapeshifter" that could become tall, wide, small, or large~\cite{cooney2017exploring}. 
Thus, various fantastical robots have been created; it has however been interestingly noted that virtual agents could be more "fantastic" than physical robots, probably due to not having a body~\cite{powers2007comparing}.

\subsubsection{Objects}
"Magic" objects have also been proposed, including gloves, cards, carpets, toys, plates, and hats:
For example, a "magic glove" has been developed that allows a human to manipulate objects with minimal force, by transmitting data to a helper robot that compensates~\cite{kazerooni2004magic}.
Additionally, "magic cards" (tags) have been proposed to command home cleaner robots, based on fairy tales of gnomes that worked when people were asleep or away; the system's ease-of-use and convenience were intended to lead to "magical experiences"~\cite{zhao2009magic}.
Furthermore, sheet-like robots have been described as "magic carpets"~\cite{kano2012sheetbot}.
Another study has proposed animating small stuffed toys from a "fantasy world (to) live with us and enchant our daily lives", where the legs and arms could be kept soft by using strings attached to an internal winch~\cite{yamashita2012stuffed}; likewise, robots with teddy bear-like features like RIBA and Moffuly have been created to carry or hug patients~\cite{mukai2010development,shiomi2017hug}.
Moreover, a small teddy bear robot with a "magic plate" has been proposed to help children to eat in a healthy manner; the plate senses eating patterns and reacts via LEDs, while the bear makes suggestions~\cite{randall2018health}.
As well, a magic hat was built, based on the "Sorting Hat" from the 
Harry Potter series,\footnote{en.wikipedia.org/wiki/Harry\_Potter} to infer emotions (happiness, sadness, and anger) in an autistic child via an EEG and communicate them via actuators~\cite{gammanpila2020sorting}.

\subsubsection{Settings}
Furthermore, some work has looked into creating magic environments, some of which integrate creatures and objects.
For example, a "magic room" was set up with projectors, lights, bubble makers, and aroma emitters, to create a "sense of magic" in providing multimodal stimulus to children with neurodevelopmental disorders ~\cite{garzotto2018magic}.
Also, a "blended reality" agent, Alphabot, composed of a box-like physical robot and a cyberphysical twin in a projected fantasy environment, was designed to move seamlessly between virtual and physical worlds~\cite{robert2012blended}; this work also described a previous effort involving a magic wand, and relates to extended or mixed reality and migratable agents (e.g., like an agent that could move between a robot and table lamp~\cite{ogawa2008itaco}).

\subsubsection{Behaviors}
Magic has been used as a kind of "oil" to enable interactions that otherwise might be impossible; e.g., in the context of teaching languages, a NAO robot "magically" waved its hand over a screen, causing a complex action to be carried out in virtual space~\cite{vogt2019second}.
Magic has also been used as a way to motivate children: During speech therapy, children practiced saying "magic words" to a robot, which opened doors and triggered various effects to move a story along~\cite{ramamurthy2018buddy}.
Fantastical robots have also been proposed as a way to encourage children to read, such as a talking owl robot that can fold around a child to create a magical private space and discuss books with them~\cite{cai2021alice}.
However, behaviors should not be completely implausible: Given that current robots can be "less attractive and friendly than a kitten, less interesting than a person, and less usable than a computer", "movie magic" suggests that designers could seek to build "character robots" that appear magical, move plausibly, and have clearly recognizable behaviors and personalities~\cite{scherer2014movie}.

\subsubsection{Potential concerns}
How people perceive such robots is likely to vary; for example, the degree to which people identify with fantastical characters has been used to quantify empathy via the Interpersonal Reactivity Index (IRI)~\cite{davis1983measuring,hsieh2022role}.
As well, a potential divide could exist between roboticists who have knowledge of what is currently possible and seek transparency--to understand underlying mechanisms in robots to provide stable solutions--and the public, who are informed by science fiction and might want to leave more room for fantasy~\cite{cheon2018futuristic}.

\subsection{Magic Tricks}

Studies have also been conducted that involve clear performances or use the "Wizard of Oz" technique.
Magic tricks have been described as a fertile ground for innovation: For example, one study explored rapid object manipulation, in introducing a robot that can do magic card tricks, which was twice as fast as a human magician at taking out the first or second card from the top of a deck of cards~\cite{koretake2015robot}.
Furthermore, a magician's gaze toward their hands when manipulating an object was found to result in increased attention, which could be applied to a robot~\cite{tamura2016human}.
Magic can also be used as a context to develop systems that can detect deception; an ability to detect human lies via pupil dilation was developed using an eye-tracker, for a magic trick by a NAO robot~\cite{pasquali2021magic}.
Likewise, robot magic tricks could be used even to promote critical thinking in humans, in addition to entertaining
~\cite{lupetti2022promoting}.

\subsubsection{Performances}
One highly entertaining magic show with a modified Baxter robot was created as a joint work between roboticists and a professional magician, Marco Tempest (a YouTube video has been viewed over 2M times and garnered 35K likes as of 2022/4/1); the accompanying study discussed timing concerns and mechanisms to alert a magician about when a robot is about to move or to allow pauses for applause~\cite{nunez2014initial}.
The magician Simon Pierro has also used a Pepper robot to perform fun illusions.\footnote{youtube.com/watch?v=zY6bG4PYVlo} 
Another important development was the creation of the Humanoid Application Challenge (HAC) Robot Magic Competition in 2016.
Various robots have been developed to perform magic tricks for this competition, including an adapted DARwIn robot that performed a magic card trick~\cite{morris2018robot}, and a robot magician called Robinion, which uses 41 DOFs, and deep learning for object recognition, to perform various tricks~\cite{jeong2022robot}.
The authors also reported that their method for generating various robot personalities resulted in increased immersion and appreciation from the audience~\cite{jeong2020robot}.

\subsubsection{Wizard of Oz}
Another common technique in HRI that could be considered to be related to magic tricks is the "Wizard of Oz" technique, where the "Wizard" in the name clearly relates to magic. For example, could manually driven vehicles in Wizard of Oz experiments be seen as AVs performing "magic" (in the sense of super-intelligent driving performance) to the passengers~\cite{yusof2016exploration,elbanhawi2015passenger}?

\subsubsection{Potential concerns}
At the same time, various work has described possible dangers of deceiving people into thinking that robots are like people.
One seminal paper paints a background for this problem, listing examples of deceptive magic devices in history, and how early misinformation might have shaped people's attitudes of how robots should look like~\cite{sharkey2006artificial}.
Likewise, there has been a historical bias to consider tricks intended to provoke wonder, like thunder and fire effects for theaters created by Hero of Alexandria, as trivial and secondary compared to more clearly practical innovations~\cite{tybjerg2003wonder}.
This leads to a question if all interactive robotics is a trivial "trick", and if robot designers are being unethical.
One response can be seen in a study that questions the need to impose a dualistic view of truth or falsehood, suggesting that robot interactions can be considered like magic performances by human magicians~\cite{coeckelbergh2018describe}.
We agree with this basic line of thought:
Designers should not need to stop prototyping designs, at the risk that some in the general public might not understand perfectly that a prototype is not a finished product.
But, roboticists should seek to be transparent about the level of technological readiness in their creations, also as standards become increasingly adopted (e.g., SOTIF (ISO/PAS 21448) for AVs, to characterize classification systems that are inherently imperfect).

Thus, various insight was gained from the literature, with references to magic found here and there. However, there seemed to be a gap: little work seems to have explored how the concept of magic could inform the design of robot interactions, which this paper seeks to address.


\section{SPECULATIVE IDEAS FOR MAGICAL ROBOT DESIGNS}
\label{section:ideation}

Inspired by the three themes explored in the literature, a speculative approach based on rapid ideation was used to explore the design space, generating some specific ideas intended to stimulate thought:

\subsection{Experiential magic}

Outside of play and affection, our ideation suggested some other considerations for supporting experiences that are fantastic, special, exciting, wonderful, easy, and liked.

The word "fantastic" means extremely good, or imaginary\footnote{dictionary.cambridge.org/ja/dictionary/english/fantastic} from its connection to "fantasy".
We believe that insight could be drawn from Symbolic Convergence Theory (SCT), which posits that the sharing of fantasies can bring people together, and can be used to gain insight into the success or failures of communications~\cite{bormann2001three}. This convergence involves a joint suspension of disbelief that can support common visions. Furthermore, experiences that also feature supernatural or illusory elements, which involve fantasy, could help support a feeling that an interaction is fantastic. 

Additionally, we considered that "special" interactions could be realized using personalization, randomness, or ephemeral components, or by coming closer to the impossible. While stereotyping can be perceived negatively (e.g., pinkification), personalization at the individual level is a common element in HRI, which can go beyond merely using a person's name, to also leverage knowledge of very frequent and infrequent behaviors enacted by a person~\cite{glas2017personal}. Some randomness balanced with a stable component (e.g., in a balance of exogenous and endogenous concerns) could act to create interest~\cite{cooney2021robot}. 
Ephemerality could be created via robots that crumble or fade over time~\cite{luria2020destruction}, or otherwise change (e.g., aging); however, the danger of corruption, which occurred with Microsoft's Tay, should be considered~\cite{chai2020keep}.
"Excitement" could be supported via emotional contagion, by showing that a robot is excited, with quick movements, aroused speech~\cite{bhimavarapu2020discriminating}, and slightly increased duration and force in touches.
"Ease" of use is a part of usability (ISO 9241-11), which itself is considered in user experience (ISO 9241-210).
Our previous work suggested that "liking" can be encouraged through displays of sincere liking by a robot with child-like appearance qualities~\cite{cooney2014affectionate}.

\subsection{Supernatural magic}

\begin{figure}[t]
\includegraphics[width=.5\textwidth]{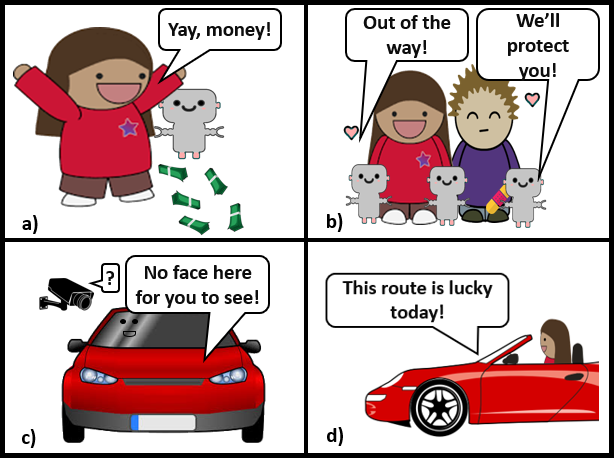}
\caption{Some "provocatype"-like ideas about how magic could be incorporated into interactions with robots: a) a HA surprises its owner with a present (money it has earned helping others in its spare time), b) HAs accompany, entertain, and protect a couple on a date, c) an AV projects a fake pattern of a face on its window to defeat unlawful face recognition technology, d) an AV tells its passenger's fortune} \label{fig_ideas}
\end{figure}

Brainstorming also led to some ideas for how robot design could be informed by fiction by emulating magical creatures, objects and powers, shown in Fig.~\ref{fig_ideas}

\subsubsection{Creatures}
Inspiration could be drawn from legends.
For example, a wise old witch robot with knowledge of nature could give advice, or a trickster or "senex" wizard robot could act as an enemy in a game for entertainment, following Jungian archetypes~\cite{mills2010children}. 
Accounts also exist of magical robot-like beings conjured up by Greek Gods and Jewish sorcerers for assistance with tasks, protection or love (e.g., humanoids of gold, bronze, stone, and earth, as well as self-moving tables)~\cite{andrist2014robots}. 
There are also numerous rarer stories of supernatural beings that could inform robot designs, from the confused Dijiang that likes to dance and sing, the Chouyu that falls asleep when it sees people, the barn-hating, pancake-loving Ovnnik, and the unhappy, ugly Squonk.\footnote{atlasobscura.com/articles/mythological-beasts-illustrations}

Insight could also be gained from popular depictions of robots in modern fiction (which could be identified, e.g., using movie database rankings), although here too differences can be expected to exist between fictional and real robots~\cite{sandoval2014human}. 
For instance, the child android Astro Boy (1952) could fly,\footnote{en.wikipedia.org/wiki/Astro\_Boy} the cat robot Doraemon had 4D pockets filled with various fantastical devices like a "portal to anywhere" (1970),\footnote{en.wikipedia.org/wiki/Doraemon} RoboCop featured a resurrected human in a cyborg suit (1987),\footnote{en.wikipedia.org/wiki/RoboCop} and a robot in Terminator 2 could repair itself in a virtually invincible manner (1991).\footnote{en.wikipedia.org/wiki/Terminator\_2:\_Judgment\_Day} (Although such works are typically presented as if their "magic" elements are realized through science, there are also some science fantasy works that combine both supernatural magic and science: e.g., Robert Heinlein's Magic Inc. (1940),\footnote{en.wikipedia.org/wiki/Magic,\_Inc.} or Godzilla vs. Mechagodzilla (1974).\footnote{en.wikipedia.org/wiki/Godzilla\_vs.\_Mechagodzilla})

\subsubsection{Objects}
Robot designs could also be informed by considering stories of magic objects (of which there are many): e.g., the Philosopher's stone, Elixir of Life, apotropaic (protective) amulets, garments of invisibility, love potions, curse effigies, magic brooms, and crystal balls~\cite{bane2020encyclopedia}.

The Philosopher's stone, said to turn base metals into gold and crystals to diamonds, inter alia, has been partly realized in physics, by creating radioactive isotopes of gold in particle accelerators.\footnote{discovermagazine.com/the-sciences/turning-lead-into-gold}
More practically, however, one idea for a magic experience could be to receive a surprise gift, e.g., of money that a robot earned when it was not completing tasks for its owner; this relates to ideas of robots as services rather than products, and a report of a Tesla owner who was using his car to mine cryptocurrencies.\footnote{cnbc.com/2022/01/08/tesla-owner-mines-bitcoin-ethereum-with-his-car.html} 
Another idea was that robots could bring magic into our environments by artistically processing and fixing objects around them; e.g., by preparing decorations, such as flower arrangements, or food such as carving fruits or 3D-printing pancakes or chocolates in fun shapes, or performing maintenance or repairs in an attractive way.\footnote{en.wikipedia.org/wiki/Kintsugi}

Regarding the Elixir of Life and Fountain of Youth, said to grant eternal youth and cure disease, robots could help people to mitigate senescence and illnesses \emph{from within}, act as vesicles for \emph{digital immortality/mind uploading}, or provide various other kinds of care:
Nanorobots could one day act as a doctor one can swallow or as drinkable magic bullets, as Ehrlich and Feynman put it; one interesting related area is liquid robotics~\cite{zhang2022reconfigurable}.
Robots have already been used to seek to preserve people's appearances and behaviors after death (e.g., in the case of Beicho Katsura III, a comic storyteller\footnote{cnet.com/culture/elderly-storytelling-android-debuts-in-japan}).
Robots also already contribute to healthcare in many ways, from surgery, to rehabilitation, helping with activities of daily living, cleaning, transporting medicine or laundry, safely treating contagious patients, and simulating patients~\cite{riek2017healthcare}.
One area in which robots could provide some "magical" assistance might be in first aid~\cite{cooney2021robotfirstaid} by conducting tasks not possible for humans, e.g., by cauterizing wounds via heat or inducing swelling via cold (e.g., with a Peltier element). Another fantastical idea considered was if a future robot could potentially be transformed into liquid for easy carrying or storage, then solidified again when needed to perform tasks.

As well, reminiscent of apotropaic magic items, Anouk Wipprecht built dresses that release smoke to hide its wearer or extend menacing spider-like legs if a potential assailant comes too close~\cite{lamontagne2017wearables}. Such robotic clothing could be adapted to incorporate more advanced inference of danger, potentially with radar to see behind walls, and defensive deterrents like OC spray, or to be smaller, in a form like a necklace or earrings. Moreover, although security robots are common in the military and home defense, bodyguard robots could be designed to accompany people.
As well, robotic devices that resemble cloaks or caps of invisibility are being designed, related to \emph{active camouflage}, \emph{optical camouflage}, and \emph{cloaking}.
One early prototype used a rear-facing camera and projector~\cite{inami2003optical}.
Likewise, various tools have been designed that could mitigate violations of privacy due to face recognition (e.g., Adam Harvey's CV Dazzle, Stealthware, and Hyperface\footnote{ahprojects.com}; or Jing Cai-Liu's Digital face projector\footnote{dailymail.co.uk/sciencetech/article-7547413/Student-project-places-one-persons-face-thwart-facial-recognition-software.html}).
One idea that emerged was that a robot could also seek to "magically" hide a person (e.g., an AV could project patterns onto its windows to skew results, if simply blackening might seem suspicious), or alternatively \emph{add} extra information and credentials to avoid any potential misunderstandings.
Another idea could to be to extend our previous work, which reported on building a robot prototype that could itself become transparent for enhanced visibility when a user's attention is not required~\cite{cooney2017exploring}, as in stories of unseen gnomes secretly helping people in need.

Regarding love potions, a robot could help to bring singles together, offering flowers, chocolates, umbrellas, rides, messages and invitations. Could a love robot somehow help to rekindle relationships, or be fashioned in a novel way?--For example, as a swarm of robots giving a couple privacy or clearing their way; or, could an AV support romance or relationship-building, and if so, could appropriate usage be ensured (age, consent, etc.)?
One disturbing thought was if conversely, a "hate robot" or "curse robot" could be built; e.g., a robot that would harass or stalk someone based on another person's wishes, possibly even committing a crime. Robot designers should take such concerns into account, and build in safeguards to avoid misuse. From the perspective of imitative magic, however, a robot could potentially provide positive experiences by mimicking an aggressor to train a person to defend themselves or relieve stress.

Moreover, like accounts of flying broomsticks or "walking on water", robots such as drones or flying AVs could enable flexible transport (e.g., like a flying motorcycle\footnote{bbc.com/news/business-60333565}).
Also, like a crystal ball, robots could seek to show a person what the world might look like in the future (e.g., using a long-term simulation, or predicting just a few seconds into the future via deep learning).
Robots could also tell fortunes for entertainment, e.g., by collating various online data, which is currently done manually by people, and displaying it through extended reality approaches, like with Google Lens\footnote{lens.google} or Microsoft HoloLens\footnote{microsoft.com/en-us/hololens}.

\subsection{Magic tricks}

Various kinds of magic exist, including vanishes, transformations, penetrations, restorations, objects or people appearing in impossible locations, predictions, and mentalism~\cite{landman2018academic}.
However, there seems to be a divide between magicians, who wish to avoid revealing how their tricks were achieved, and roboticists, who wish to share information transparently to enable progress in the field. In our ideation, we respected the magician's perspective above, and thus avoid discussing possible tricks here in this section.

\section{PRACTICAL IMPLEMENTATION}
\label{section:implementation}

The real world often raises new problems.
Therefore, some of our theoretical ideas from the previous section were applied to build a prototype of a robot that can do magic tricks. Ideas from all three themes of the paper were included.

For the "experiential magic" component, we gave our robot, a Baxter robot, a playful appearance and behaviors: the robot was equipped with a magician's hat, wand, bow tie, and cape--out of place on an obviously mechanical body--as well as some dramatic wand motions and sounds like a deep ominous laugh. We also sought to incorporate humor into the interactive concept, for which we selected the idea of a curious human trying to figure out if a reluctant magician robot had magic powers or not; for each successful trick, the robot was to also do something doubtful. 
To make it seem like the robot was able to interact contingently and in a personal way, we came up with a predetermined script for the robot's utterances. To reduce the risk of failures, speech recognition was used sparingly; instead, known timings and touches to the robot's arms were used to move the script along.
To make the interaction feel special, we decided that the main magic tricks should be new. The unique appearance of the robot--obviously a machine but clothed with a magician's attire--was also intended to contribute to this effect.

For the supernatural magic component, we focused on the concept of "mentalism", that our robot could pretend to have psychic powers, like the ancient fortune-tellers and crystal balls of legend. To realize this concept, we started from the idea that sight is a powerful modality, and that people tend to see the world from their own perspectives; i.e., despite the presence of x-ray machines in constrained environments like hospitals, our guess was that lay persons have little understanding of how the world appears outside of the visual light spectrum. Therefore, we decided to apply some multispectral analysis using a thermal camera and ultraviolet (UV) camera to realize a display of psychic powers for the main part of the performance.

It was desired that the robot itself would handle the tricks (a human magician should not be required) but that there should be a good amount of audience interaction (in the simplest case, with just one person). Thus, four tricks were conceived. Despite the concept of "growing implausibility" which is sometimes used in magic performances, in which tricks become increasingly complex to leave a strong impression at the end~\cite{landman2018academic}, we wanted the easier tricks to be at the start and end, to still be able to provide a good impression in case of a failure, due to the initial nature of the exploration (our team did not include a professional magician) and various time constraints. Therefore, tricks 1 and 4 were designed to be simple, with the main interest intended to be elicited in tricks 2 and 3: 
\begin{enumerate}
  \item Trick 1 involved the robot suddenly transforming a wand in its right hand to reveal a message. A simple fold was used to conceal the message, and the robot's gripper was used to display or hide it. The concealed message was playful, to leave the human in doubt about the robot's abilities.
  \item Trick 2 involved the robot seeing through seemingly opaque material. For this, we built a simple prop, a box outfitted with black plastic from a garbage bag, which is opaque in the visual spectrum but transparent in thermal. To see inside the box, the robot used a thermal camera hidden within its magician's hat. It was set to use a threshold on the standard deviation (SD) within the region of interest of the box to determine if a human hand was inside or not. SD was chosen for simplicity, since it is robust to changes in illumination (in our lab, the overhead lights often flicker and even shut off automatically, and sunlight often fell on the robot and box, which could change quickly due to cloud movements). Basically, a high SD was detected when there was a warm human hand inside the room temperature box, and low otherwise. After performing its trick successfully, the robot then again exhibited playful behavior: When the human asked the robot how many fingers he is holding up, the robot gave an obviously impossible answer, "64". 
  \item Trick 3 involved the robot seeing something "invisible": if a person had applied sunscreen to his hand or not. For this, the robot used its motions, reaching out for the human's hand, to "trick" the human into placing their palm in view of a UV camera attached to its chest. A UV flashlight attached to the robot's left hand (hidden by a wand) was also used to shine UV light onto the person's hand to better recognize (since the setting was indoors; if outside, this would not be needed). During development, just in case, UV goggles were used for eye protection. Then, as in the trick before, a simple empirically-determined threshold was used. If a person applied sunscreen, their hand would appear mostly dark in the UV spectrum (due to absorbing the UV light); else their hand would appear light.
  \item Trick 4 involved a common mind-reading trick based on speech recognition. The robot asked the person to think of a color (red, green, or blue), thus limiting the challenge to a few options. The robot then asked the person which color they had chosen. If a response from the person was detected, the robot asked the person to check in a nearby location revealing, to the person's surprise, a sheet of paper in the color the person had been thinking of. The core of the trick was that a sheet of each color was hidden in a different location around the robot, so it only had to change its message based on what it heard.
\end{enumerate}

\begin{figure}[t]
\includegraphics[width=.5\textwidth]{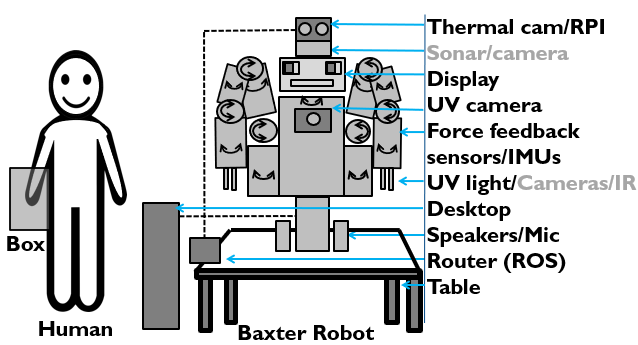}
\caption{The system set-up (components in gray were not used in the tricks).} \label{fig_setup}
\end{figure}

\begin{figure}[t]
\includegraphics[width=.5\textwidth]{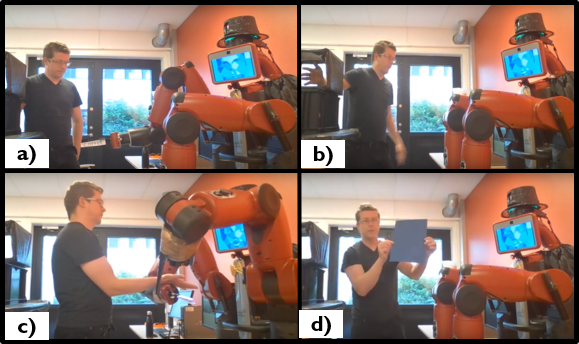}
\caption{Scenes from the magic performance: a) the robot opens its "wand" to reveal a message, b) the robot sees through a black box using a thermal camera, c) the robot detects sunscreen on a hand via an UV camera, and d) the robot pretends it knew which color a human was thinking of using a standard "cold reading"/"rainbow ruse"-style magic trick.} \label{fig_scenes}
\end{figure}

Some technical challenges involved robust recognition in the presence of varying light conditions, reflections, and noise, inexpensive components that sometimes need to be adjusted and restarted, and setting up the ROS communication between the three computers required to run the robot and sensors. The hardware and software used is briefly described in Table~\ref{tab_tools}, main hardware components are shown in Fig.~\ref{fig_setup}, and scenes from a trial run are shown in Fig.~\ref{fig_scenes}. A video of a trial run can also be seen online.\footnote{youtube.com/watch?v=Ew8X3oqre14}

\begin{table}
\caption{Software and Hardware.}
\label{tab_tools}  
\begin{tabularx}{\linewidth}{ | >{\hsize=0.5\hsize}X | >{\hsize=0.5\hsize}X |  } \hline
Component           & Tool  \\ \hline
Robot           & Baxter  \\ \hline
Image processing    & OpenCV  \\ \hline
Operating Systems   & Ubuntu, Raspbian  \\ \hline
Speech recognition/generation  & Pocketsphinx, Festival  \\ \hline
Communication           & Robot Operating System (ROS)  \\ \hline
Thermal-visual sensing           & FLIR  80 x 60 , 8-14µm, RGB  \\ \hline
UV           & R1080P (up to 3264 x 2448), 320nm-380nm  \\ \hline
\end{tabularx} 
\end{table}

Our system was presented as part of the Human Application Challenge (HAC) Robot Magic and Music Competition.\footnote{facebook.com/humanoidchallenge}
A master's student was also involved to assist in development and performances, also with the goal of sharing the experience and inspiring our next students. 
Of 12 entrants from five countries (Taiwan, Iran, Brazil, Canada, and Sweden), eight teams went on to a final round held live on August 23, 2021.
The final task called for a 20 minute magic performance and slide presentation, which was judged on technical difficulty, novelty, and showmanship.
The judges for the competition comprised seven experts in AI and robotics from universities and companies, as well as performing artists, from six countries: Korea, Portugal, Malaysia, Taiwan, Japan, and Canada.
Some troubles were encountered: During the performances, we had planned to watch the other teams perform and practice our performance one more time just before our turn, but were prevented from doing so when the team just before us ended ahead of schedule; thus, we weren't able to detect the UV flashlight's battery dying just as we were performing the sunscreen trick, giving a faulty reading. (Accessing the battery was also tricky and time-consuming due to how it was mounted, so we adjusted our trick and proceeded.)
As well, the international nature of the competition meant that many of the judges and teams were up in the middle of the night; this might have made it more difficult to follow along with our storyline, which contained much speech, not all of which might have been required.
As well, many of the other teams had competed in previous years and had developed very enjoyable performances.
Nonetheless, our system received a positive appraisal from the judges, placing 2nd in the competition, with the prize of a small robot gripper.\footnote{seedrobotics.com} We believe this provided some additional verification for our design ideas.

\section{DISCUSSION}
\label{section:discussion}

The contribution of the current paper was exploring the concept of "magic" in HRI:
\begin{itemize}
\item{\emph {Theory}. We presented the results of a rapid, scoping review to bring together various work relating to magic and fantasy that has been done in HRI, grouping studies into three themes related to experiential magic, supernatural magic, and illusory magic. This was extended by presenting some of our own ideas about how magic could be integrated into various designs.}
\item{\emph {Practice}. We also described a simplified prototype built based on applying some of our ideas, which received an award in an international competition, also mentioning some challenges that arose.}
\end{itemize}

The current study is limited by the complexity of the phenomenon (numerous aspects and definitions of magic had to be considered), the rapid scoping nature of the review and secrecy around magic tricks (various related work no doubt exists that has not been identified here), the cursory nature of our ideation, and the simplified prototype we built (more advanced prototypes could use Transformers or GANs for recognition or anticipation of human actions).

Nonetheless, our aim has been to shine light on an intriguing phenomenon that seems relevant to various robot designs but has not yet received much attention in HRI--to raise awareness and stimulate discussion about how the robots of tomorrow might look. As Solange Nicole and Drummond Money-Coutts put it, "The world will always need magic" and "Magic touches people in the way great art does. It lets them see the world with new eyes"\footnote{everydaypower.com/magic-quotes}--an important ability in today's world. 

\addtolength{\textheight}{-2cm}   




\section*{ACKNOWLEDGMENT}

We thank our master's student, Vivek Uddagiri, as well as the organizer, Jacky Baltes, judges, and competitors in the HAC Robot Magic and Music competition, for a "magic" experience.



\bibliographystyle{IEEEtran}
\bibliography{IEEEabrv,magic}

\end{document}